\documentclass[10pt,twocolumn,letterpaper]{article}
\usepackage{authblk}
\makeatletter
\renewcommand\AB@affilsep{\quad}
\makeatother
\usepackage[pagenumbers]{iccv} 

\definecolor{iccvblue}{rgb}{0.21,0.49,0.74}
\usepackage[pagebackref,breaklinks,colorlinks,allcolors=iccvblue]{hyperref}
\usepackage{multirow}
\usepackage{array}
\usepackage{graphicx}
\usepackage{subcaption}
\usepackage{makecell}

\def\paperID{4884} 
\def\confName{ICCV}
\def\confYear{2025}

\title{V-STaR: Benchmarking Video-LLMs on \underline{V}ideo \underline{S}patio-\underline{T}empor\underline{a}l \underline{R}easoning}

\author{\vspace{-25pt}Zixu Cheng\textsuperscript{1}, Jian Hu\textsuperscript{1}\thanks{corresponding author}, Ziquan Liu\textsuperscript{1}, Chenyang Si\textsuperscript{2}, Wei Li\textsuperscript{3}, Shaogang Gong\textsuperscript{1}\\\vspace{-10pt}
\textsuperscript{1}Queen Mary University of London,
\textsuperscript{2}Nanjing University, \textsuperscript{3}Nanyang Technological University\\\vspace{-3pt}
{\tt\small \{zixu.cheng,jian.hu,ziquan.liu,s.gong\}@qmul.ac.uk,chenyang.si@nju.edu.cn,wei.l@ntu.edu.sg} \\ \textcolor{purple!80}{https://V-STaR-Bench.github.io/} \vspace{-10pt}}

\begin{document}
\twocolumn[{
    \renewcommand\twocolumn[1][]{#1}
    \maketitle
    \vspace*{-0.2in}
    \centering
    \captionsetup{type=figure}
    \vspace{-0.2cm}
    \includegraphics[width=1\textwidth]{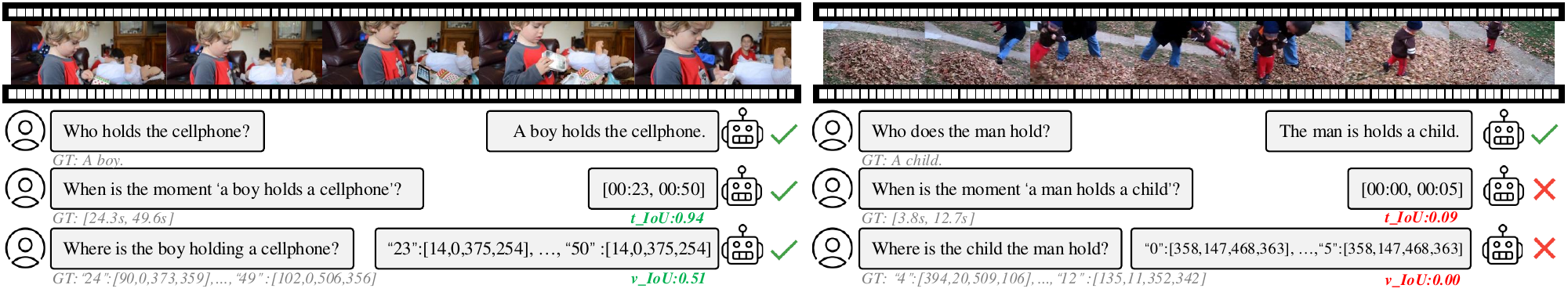}
    \vspace{-0.7cm}
    \caption{Illustration of the challenge in evaluating spatio-temporal reasoning ability. 
   In both examples, the model correctly identifies objects, but its performance on related temporal and spatial questions varies greatly. 
   This inconsistency suggests that correct answers may result from pretraining co-occurrence biases rather than true understanding. Existing benchmarks focus on object identification but fail to determine whether models truly engage in spatio-temporal reasoning. Our V-STaR benchmark fills this gap by evaluating how models integrate spatial, temporal, and causal relationships in video understanding.}
    \label{fig:motivation}
    \vspace{0.3cm}
}]

\begin{abstract}

\vspace{-10pt}
\noindent Human processes video reasoning in a sequential spatio-temporal reasoning logic, we first identify the relevant frames (``when") and then analyse the spatial relationships (``where") between key objects, and finally leverage these relationships to draw inferences (``what").
However, can Video Large Language Models (Video-LLMs) also ``reason through a sequential spatio-temporal logic” in videos?
Existing Video-LLM benchmarks primarily focus on assessing object presence, neglecting relational reasoning. Consequently, it is difficult to measure whether a model truly comprehends object interactions (actions/events) in videos or merely relies on pre-trained ``memory" of co-occurrences as biases in generating answers. In this work, we introduce a \textbf{V}ideo \textbf{S}patio-\textbf{T}empor\textbf{a}l \textbf{R}easoning (V-STaR) benchmark to address these shortcomings. The key idea is to decompose video understanding into a Reverse Spatio-Temporal Reasoning (RSTR) task that simultaneously evaluates what objects are present, when events occur, and where they are located while capturing the underlying Chain-of-thought (CoT) logic. To support this evaluation, we construct a dataset to elicit the spatial-temporal reasoning process of Video-LLMs. It contains coarse-to-fine CoT questions generated by a semi-automated GPT-4-powered pipeline, embedding explicit reasoning chains to mimic human cognition. 
Experiments from 14 Video-LLMs on our V-STaR reveal significant gaps between current Video-LLMs and the needs for robust and consistent spatio-temporal reasoning. 

\renewcommand{\thefootnote}{\fnsymbol{footnote}}
\footnotetext[1]{Corresponding author.}

\end{abstract}    

\vspace{-20pt}
\section{Introduction}
\label{sec:intro}
When answering a video question, humans first identify the relevant moment (\textit{``when"}), then establish spatial and temporal relationships (\textit{``where"}-\textit{``when"} dependencies) of the key objects. Finally, we use these relationships to infer the answer (\textit{``what"})~\cite{li2022representation}.
%
This reflects humans' natural ability to construct a sequential spatio-temporal reasoning logic by progressively organizing events across time and space~\cite{signorelli2020cognitive}.

This structured reasoning process has inspired the AI community’s development of Chain-of-Thought (CoT) reasoning~\cite{lanham2023measuring,sprague2024cot,guo2025deepseek}.
Due to the inherent sequential logic of language, CoT can not only enhance model performance but also serve as a tool for evaluating the reasoning capabilities of LLMs.
However, unlike text-based tasks, visual tasks often lack clear logical steps, making it more challenging to design effective CoT strategies for both reasoning training and evaluation~\cite{chen2023measuring}.
This challenge is further compounded in video reasoning, where understanding requires not only recognizing objects (\textit{``what"}) but also establishing their spatial (\textit{``where"}) and temporal (\textit{``when"}) relationships.

\begin{table}[tb]
\setlength{\tabcolsep}{0.7pt} 
\centering
\scalebox{0.58}{
\begin{tabular}{l|c|>{\centering\arraybackslash}p{1.3cm}|>{\centering\arraybackslash}p{1.4cm}|>{\centering\arraybackslash}p{1.3cm}|c|c|c}
\hline
\multirow{2}{*}{Benchmark} & \multirow{2}{*}{Venue}  & \multicolumn{3}{c|}{VQA with Grounding}  & \multirow{2}{*}{\makecell{CoT\\Questions}} & \multirow{2}{*}{Tasks} & \multirow{2}{*}{Metrics} \\
\cline{3-5}
& & VQA & Temporal & Spatial & & & \\
\hline
VidSTG~\cite{zhang2020does} & CVPR'20 & - & \checkmark & \checkmark & - & Grounding & Rule-based \\
HC-STVG~\cite{tang2021human} & TCSVT'22 & - & \checkmark & \checkmark & - & Grounding & Rule-based \\
MVBench~\cite{li2024mvbench} & CVPR'24 & \checkmark & - & -  & - & MCQ & Accuracy \\
VideoMME~\cite{fu2024video} & CVPR'25 & \checkmark & - & -  & - & MCQ & Accuracy \\
TempCompass~\cite{liu2024tempcompass} & ACL'24 & \checkmark & - & -  & - & MCQ or Y/N & Accuracy \\
Movie-Chat-1k~\cite{song2024moviechat} & CVPR'24 & \checkmark & - & -  & - & Open-ended & LLM-based \\
MMBench-Video~\cite{fang2025mmbench} & NeurIPS'24 & \checkmark & - & -  & - & Open-ended & LLM-based \\
LongVideoBench~\cite{wu2025longvideobench} & NeurIPS'24 & \checkmark & - & - & - & MCQ & Accuracy \\
HourVideo~\cite{chandrasegaran2025hourvideo} & NeurIPS'24 & \checkmark & - & -  & - & MCQ & Accuracy \\
TVQA~\cite{lei2018tvqa} & EMNLP'18 & \checkmark & \checkmark & -  & - & Open-ended & Rule-based \\
QAEgo4D~\cite{barmann2022did} & CVPRW'22 & \checkmark & \checkmark & - & - & Open-ended & Rule-based \\
NeXT-GQA~\cite{xiao2024can} & CVPR'24 & \checkmark & \checkmark & -  & - & Open-ended & Rule-based \\
REXTIME~\cite{chen2025rextime} & NeurIPS'24 & \checkmark & \checkmark & - & - & Open-ended & Rule-based \\
E.T. Bench~\cite{liu2024bench} & NeurIPS'24 & \checkmark & \checkmark & - & - & Open-ended & Rule-based \\
GCG~\cite{munasinghe2024videoglamm} & ArXiv'24 & \checkmark  & - & \checkmark & - & Open-ended & Rule-based \\
TVQA+~\cite{lei2020tvqa+} & ACL'20 & \checkmark & \checkmark & \checkmark  & - & Open-ended & Rule-based \\
\hline
Ours & - & \checkmark & \checkmark & \checkmark  & \checkmark & Open-ended & Rule-based \\
\hline
\end{tabular}
}
\vspace{-8pt}
\caption{Comparison of spatial-temporal understanding datasets.}
\vspace{-15pt}
\label{tab:benchmark-comparison}
\end{table}

%
Some video spatio-temporal benchmarks attempt to evaluate models' reasoning abilities.
But current benchmarks only measure models' output on object names (answering \textit{``what"}) without assessing models' capacity for relational reasoning. 
As a result, as shown in Fig.~\ref{fig:motivation}, models can achieve high accuracy in question answering tasks by leveraging pre-trained co-occurrence biases~\cite{hu2025leveraging} rather than truly understanding object interactions spatio-temporally. We argue it is essential to quantify a model’s spatio-temporal reasoning ability. This helps reveal a Video-LLM’s true limitations and potential in video understanding tasks.
%

However, existing datasets lack a structured framework to assess spatio-temporal reasoning ability. As shown in Tab.\ref{tab:benchmark-comparison}, numerous datasets ~\cite{li2024mvbench,chen2025rextime,liu2024bench,wu2025longvideobench,lei2018tvqa} typically focus on three aspects: \emph{``what"} 
objects are present, \emph{``when"} events occur, and \emph{``where"} objects are located. However, they either cover only one aspect or treat such questions in isolation as separate sub-tasks~\cite{lei2020tvqa+, li2024videovista}, failing to measure models' ability of logical spatio-temporal reasoning. 
Effective video understanding requires integrating \emph{``what"}, \emph{``when"}, and \emph{``where"} through CoT-style reasoning.

In this work, we introduce a new Video Spatio-Temporal Reasoning (V-STaR) benchmark, to evaluate explicitly the capacity of current Video-LLMs on spatio-temporal reasoning comprehensively. 
There are two distinct designs in our V-STaR.
First, we propose a Reverse Spatio-Temporal Reasoning (RSTR) task, to break down and quantify a model's spatio-temporal reasoning ability. RSTR simultaneously assesses a model's output on what objects are present, when events occur, and where objects are located while also examining how a model constructs CoT logic during reasoning.
Second, to support this evaluation, we construct a fine-grained reasoning dataset using a semi-automated GPT-4-powered pipeline. To mimic a “human-thought” cognitive process, we embed explicit reasoning chains within custom \texttt{<think><think>} tags for each question. To mitigate error propagation in model reasoning, we further decompose these reasoning chains into structured CoT tasks with increasing granularity, enabling a more systematic and fine-grained assessment of spatio-temporal video understanding.
Additionally, we propose a new metric, the Logarithmic Geometric Mean (LGM), which combines model score at each step of the reasoning chain, offering a comprehensive assessment of spatio-temporal reasoning. We conducted experiments on 14 contemporary and state-of-the-art models, providing an inclusive assessment of the Video-LLMs' reasoning capabilities.
\textbf{Our contributions are as follows:}

\noindent1) We are among the first to investigate the spatio-temporal reasoning ability of state-of-the-art Video-LLMs, revealing their unreliable inference in such tasks. To support the evaluation, we propose V-STaR, the first benchmark explicitly designed to evaluate Video-LLM's spatio-temporal reasoning ability in answering questions explicitly in the context of {\em ``when", ``where"}, and {\em ``what"}.



\noindent2) We construct a fine-grained reasoning dataset with coarse-to-fine CoT questions, enabling a structured evaluation of spatio-temporal reasoning. 
Specifically, we introduce a Reverse Spatio-Temporal Reasoning (RSTR) task to quantify models' spatio-temporal reasoning ability.


\noindent3) Experiments from 14 Video-LLMs on V-STaR reveal although many models perform well on \textit{``what"}, some struggle to ground their answers in time and location. This finding highlights a fundamental weakness in existing Video-LLMs regarding causal spatio-temporal reasoning and inspires research in improving trustworthy spatio-temporal understanding in future Video-LLMs. 




\section{Related Works}
\noindent\textbf{Spatio-temporal understanding in Video-LLMs}
Video-LLMs~\cite{li2023videochat, zhang2024video, wang2024qwen2, ren2024timechat, huang2024vtimellm, guo2024trace, yuan2025sa2va} have made rapid progress in video understanding, enabling them to answer a diverse range of questions about videos, e.g. framed as video question answering (VQA) problems. 
Many open-source Video-LLMs demonstrate competitive results to the proprietary commercial models, e.g., GPT-4o~\cite{OpenAIGPT4o} and Gemini-2-Flash~\cite{Gemini2}, across multiple Video-LLM Benchmarks~\cite{fu2024video,wu2025longvideobench,fang2025mmbench}.
Recent studies have explored the ability of Video-LLMs in video temporal and spatial understanding. 
TimeChat~\cite{ren2024timechat}, VTimeLLM~\cite{ren2024timechat}, and Trace~\cite{guo2024trace} were among the first to develop specialized models for video temporal grounding, which involves localizing event timestamps in a video given a text description. 
Additionally, general-purpose models, such as Qwen2.5-VL~\cite{bai2025qwen2} and VideoLlama3~\cite{zhang2025videollama}, also exhibit strong temporal grounding capability in video, achieving comparable performance of classic models~\cite{zhang2020learning,zhang2020span} on temporal grounding datasets~\cite{gao2017tall,krishna2017dense}. 
While certain Video-LLMs~\cite{wang2024qwen2,bai2025qwen2,zhang2025videollama} claim to support object detection~\cite{lin2014microsoft} and referring expression comprehension~\cite{yu2016modeling} on image inputs, their video spatial grounding capabilities remain largely unexplored. 
\citet{munasinghe2023pg} first introduces spatial grounding to Video-LLMs, later extended to video segmentation~\cite{yan2024visa, munasinghe2024videoglamm, yuan2025sa2va}. However, most existing Video-LLMs evaluate their performance on VQA, temporal grounding, and spatial grounding tasks separately, without validating their ability for spatio-temporal reasoning. It is unclear whether Video-LLMs correctly understand and use spatio-temporal information in video reasoning.

\begin{figure*}[tb]
   \centering 
   \vspace{-20pt}
   \includegraphics[width=\textwidth]{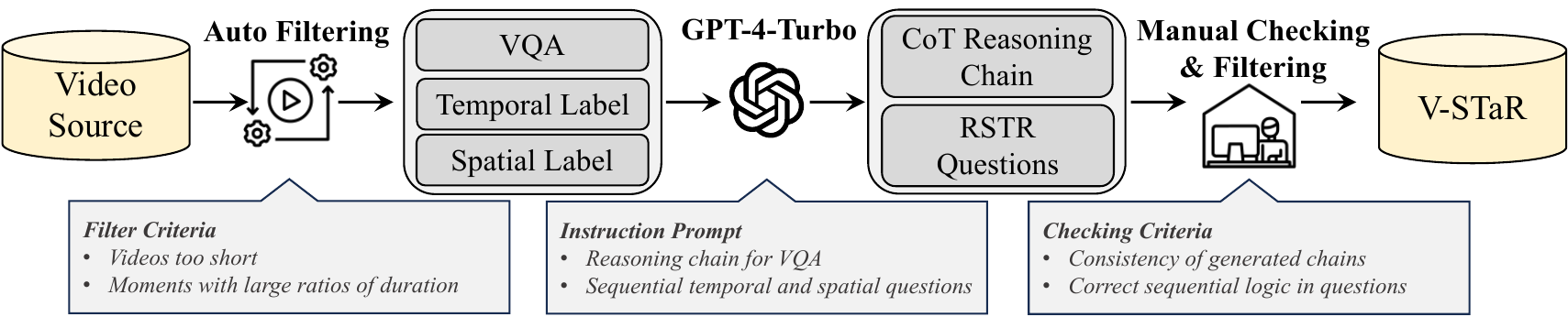}
   \vspace{-20pt}
   \caption{Illustration of the semi-automated data construction pipeline of V-STaR. GPT-4 generates a spatio-temporal reasoning CoT chain to answer VQA questions, along with a set of RSTR questions. The RSTR questions are independent temporal or spatial grounding challenges, decomposed from the CoT reasoning chain, designed to evaluate the model’s spatio-temporal reasoning capabilities.}
   \label{fig:pipeline}
\vspace{-15pt}
\end{figure*}

\noindent\textbf{Video-LLM Benchmarks}
Recently, numerous benchmarks have been proposed to evaluate the general video understanding and reasoning capabilities of Video-LLMs. These benchmarks span a diverse range of tasks~\cite{li2024mvbench, li2024videovista, liu2024tempcompass}, types~\cite{fu2024video,fang2025mmbench,song2024moviechat} and durations~\cite{wu2025longvideobench,chandrasegaran2025hourvideo,zhou2024mlvu}. 
However, they primarily focus on Video Question Answering (VQA), essentially addressing the \textit{``what"} question in videos while overlooking whether a model correctly understands and leverages spatio-temporal context in their reasoning process.
To bridge this gap, some studies~\cite{xiao2024can, chen2025rextime, munasinghe2024videoglamm} have begun incorporating temporal or spatial grounding to validate the reasoning pathways of Video-LLMs.
TVQA~\cite{lei2018tvqa} proposed Grounded Video Question Answering (GVQA), requiring models to answer not only multiple-choice questions but also temporal grounded evidence in TV series videos. 
Expanding upon GVQA, benchmarks such as QAEgo4D~\cite{barmann2022did}, Next-GQA~\cite{xiao2024can}, and ReXTime~\cite{chen2025rextime} 
have extended these tasks to ego-centric videos, real-world videos, and complex reasoning questions. 
Grounded Conversation Generation (GCG)~\cite{munasinghe2024videoglamm} was designed to challenge models in reasoning and identifying specific objects for segmentation in videos.
VidSTG~\cite{zhang2020does} further integrated spatio-temporal grounding with interrogative queries to reason the referred object in videos. TVQA+~\cite{lei2020tvqa+} then introduced spatio-temporal grounding for VQA, but treated it as three independent sub-tasks, without investigating how models utilize temporal and spatial relationships in their reasoning process. 
Building on these works, our benchmark introduces CoT reasoning and employs temporal and spatial grounding as a structured reasoning chain, aiming to explicitly investigate the spatio-temporal reasoning abilities of Video-LLMs, providing a more comprehensive evaluation framework.

\section{V-STAR Benchmark}
In this section, we first define the Reverse Spatio-Temporal Reasoning (RSTR) task for evaluating the spatio-temporal reasoning capabilities of Video-LLMs. Then, we introduce a semi-automatic pipeline using GPT-4~\cite{OpenAIGPT4}, to generate coarse-to-fine RSTR questions to construct the dataset.

\begin{figure*}[ht]
   \centering
   \vspace{-10pt}
   \includegraphics[width=\textwidth]{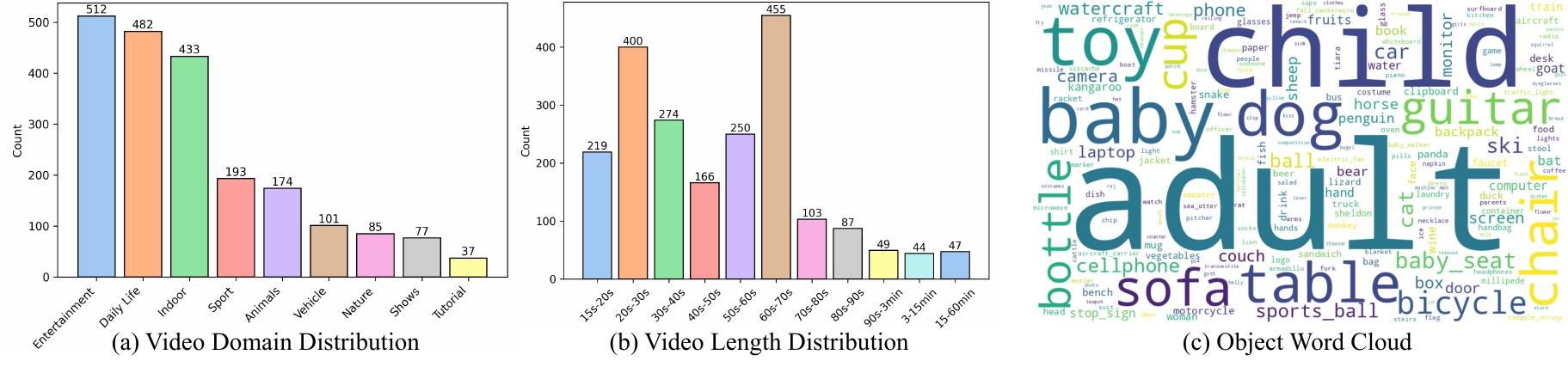}
   \vspace{-20pt}
   \caption{Dataset statistics of video domain and length, and visualization of objects in video.}
   \label{fig:statistics}
\end{figure*}

\begin{table*}[!t]
    \centering
    \renewcommand{\arraystretch}{0.7} 
    \scalebox{0.90}{
    \begin{tabular}{l|cccccccccc}
        \toprule
        & Entertainment & Daily Life & Indoor & Sports & Animals & Vehicles & Nature & Shows & Tutorial & Overall \\
        \midrule
        Avg Length(s) & 104.60 & 88.21 & 45.24 & 128.00 & 38.07 & 42.16 & 44.19 & 258.14 & 1512.05 & 110.23 \\
        Avg Moment(s) & 9.32 & 8.68 & 10.40 & 6.99 & 8.10 & 7.70 & 8.96 & 10.71 & 10.45 & 9.06\\
        Avg M/L Ratio(\%) & 15.16 & 20.29 & 22.98 & 20.76 & 21.30 & 20.01 & 20.49 & 18.34 & 2.02 & 19.32\\
        Num of BBox & 2097 & 4351 & 4621 & 1409 & 1471 & 806 & 789 & 840 & 409 & 16793\\
        Num of Objects & 255 & 38 & 29 & 37 & 26 & 16 & 12 & 18 & 29 & 342\\
        \bottomrule
    \end{tabular}}
    \vspace{-8pt}
    \caption{Statistical comparison of different domains.}
    \vspace{-8pt}
    \label{tab:category_comparison}
\end{table*}

\subsection{Task Definition}
Most existing reasoning tasks require a model to directly produce answers to complex sequential problems. These benchmarks~\cite{li2024mvbench,fu2024video} often fail to reveal the model’s underlying reasoning process, and the model may exploit pre-trained biases rather than engage in genuine reasoning on a given video.
To truly assess a model's ability, we propose the Reverse Spatio-Temporal Reasoning (RSTR) task. Specifically, the task is based on three fundamental elements: \textit{``what"}, \textit{``when"}, and \textit{``where"}. Based on the spirit of human problem-solving ~\cite{signorelli2020cognitive}, when faced with a complex video spatio-temporal reasoning challenge, people typically start by (1) identifying the relevant frames (\textit{``when"}), (2) then determining and analyzing the positions of objects in those frames (\textit{``where"}), and (3) finally answering the \textit{``what"} question. In contrast, due to pre-training co-occurrence biases, even if a Video-LLM produces a correct answer, it is hard to tell whether it did so via its own reasoning process or by relying on prior knowledge~\cite{hu2025leveraging}, causing inconsistency and sensitivity to prompting in a model’s answers, e.g. hallucinations at the wrong time in the wrong place.
To evaluate this, we adopt a Reverse CoT strategy: the model is first prompted to answer the \textit{``what"} question, and then, based on that answer, a coarse-to-fine reasoning chain following the order \textit{``what-when-where"} evaluates the model’s spatial-temporal reasoning capability. 
We also design a parallel chain in the order \textit{``what-where-when"} to examine how different logical sequences impact the final results.
Our RSTR task not only evaluates the model’s spatio-temporal reasoning ability, but also quantifies the influence of various logical sequences.

\subsection{Dataset Construction}
A significant challenge in constructing this new dataset is to obtain videos accompanied by precise, coarse-to-fine CoT questions. To ease the burden of manual annotation, we propose a hybrid approach that leverages annotated data from existing datasets while incorporating a semi-automated annotation pipeline. This approach unfolds in three stages: data collection, pipeline construction, and metric design.

\noindent\textbf{Data Collection.}
%
We collected videos from datasets that offer spatial and temporal grounding. 
We used VidSTG~\cite{zhang2020does}, TVQA+~\cite{lei2020tvqa+}, and GOT-10K~\cite{huang2019got} datasets. VidSTG provides spatio-temporal grounding. TVQA+ offers temporal grounding for certain objects through question-answer pairs. GOT-10k gives spatial grounding details. However, these datasets do not include CoT reasoning chains, and their video durations are mostly between 0 and 3 minutes. Such rather short video durations are much narrower than what is seen in real-world scenarios.
To ensure a diverse range of video durations, we started with the GOT-10k dataset because it has complete spatial grounding information. We then collected additional videos from YouTube that range from 3 minutes to 1 hour. We randomly inserted selected GoT videos into various positions within these videos. It ensures that the final dataset shows a high degree of diversity in both duration and content. Once we gathered videos of various lengths and types, we built the coarse-to-fine CoT questions for reasoning.

\noindent\textbf{Pipeline Construction.}
In the previous stage, we collected a diverse set of videos with complete spatio-temporal grounding labels. However, our goal is to evaluate the model's spatio-temporal reasoning ability in a fine-grained manner. To achieve this, we leveraged GPT-4-turbo~\cite{OpenAIGPT4} to construct a semi-automated pipeline for generating CoT reasoning chains and questions with a coarse-to-fine granularity. Specifically, as shown in Fig.~\ref{fig:pipeline}, we first automatically filter out samples where the video length is too short or the moment ratio to video length is too large, ensuring that the questions remain sufficiently challenging. Next, we input a video question along with its answer, as well as the corresponding temporal and spatial annotations, into GPT-4-turbo. This process generates a spatio-temporal reasoning chain for answering the video question, which is then decomposed into two independent fine-grained sub-questions focusing on temporal and spatial localization. These sub-questions evaluate whether the model's spatial and temporal reasoning is correct. Finally, we manually verify the reasoning chain and the decomposed localization questions, assigning the temporal and spatial labels to each sub-question. 

Furthermore, to comprehensively investigate how a model leverages temporal and spatial relationships in the reasoning, we formulate our generated questions into two RSTR task chains:\textit{``what-when-where"} and \textit{``what-where-when"}. In each reasoning chain, the subsequent question incorporates the ground truth of the previous question. For instance, in the \textit{``what-when-where"} chain, the \textit{``when"} question contains the ground truth of the \textit{``what"} question, and the ``where" question includes the ground truths of both the \textit{``when"} and \textit{``what"} questions. This design prevents the model from making errors in earlier reasoning steps and propagating to the final result, allowing for an independent and fairer evaluation of temporal and spatial reasoning. Ultimately, each sample is associated with one spatio-temporal CoT reasoning chain and two RSTR task chains.

\begin{figure*}[!th]
   \centering 
   \vspace{-12pt}
   \includegraphics[width=\textwidth]{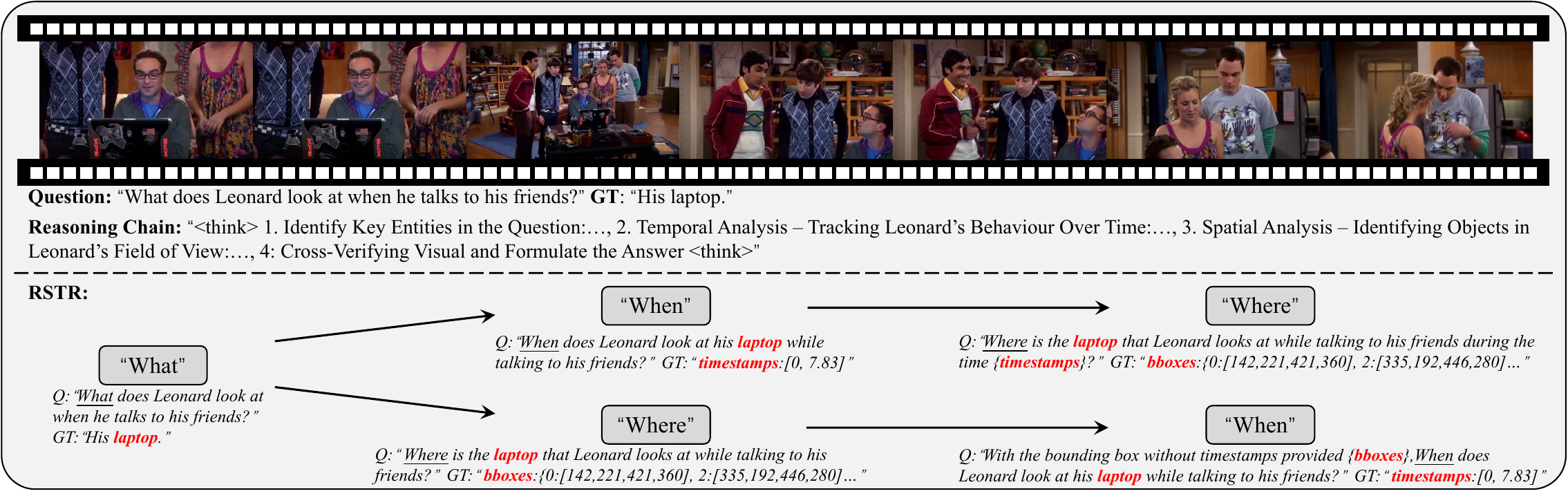}
   \vspace{-15pt}
   \caption{An example illustrating the construction of CoT questions. Each sample contains a thinking chain and two RSTR question chains.}
   \label{fig:sample}
\vspace{-5pt}
\end{figure*}

\noindent\textbf{Metric Design.}
To evaluate the model's spatio-temporal reasoning ability, we have decomposed the task into fine-grained CoT reasoning questions. Each question targets one of the \textit{``what"}, \textit{``when"}, and \textit{``where"} components, which are independently assessed using $\textit{Acc}$ (accuracy), $\text{m\_tIoU}$ (mean temporal IoU), and \(\text{m\_vIoU}\) (mean visual IoU). Although this method effectively measures the performance of each individual question, it only considers the correctness of each answer in isolation, ignoring the interconnections between different answers within CoT reasoning.  

To overcome this problem, we propose evaluating the model’s overall performance across these three questions using the Arithmetic Mean (AM) (Eq.~\ref{eq:1}) and a modified logarithmic Geometric Mean (LGM) (Eq.~\ref{eq:4}).  
Specifically, AM is given as:  \setlength{\belowdisplayskip}{1.6pt} \setlength{\belowdisplayshortskip}{1.6pt}
\setlength{\abovedisplayskip}{1.6pt} \setlength{\abovedisplayshortskip}{1.6pt}
\begin{align} 
\label{eq:1}
\small
AM = \frac{1}{3}(\textit{Acc} + \text{m\_tIoU} + \text{m\_vIoU}),
\end{align}
while AM effectively assesses the model’s overall performance across different metrics, it is susceptible to extreme values.  
To mitigate this issue, we employ the Geometric Mean (GM) to evaluate model performance:  
\begin{align} 
\label{eq:original}
\small
GM = (\textit{Acc} \times \text{m\_tIoU} \times \text{m\_vIoU})^{\frac{1}{3}},
\end{align}
However, when any of the metrics is zero, GM will become zero, which fails to reflect the contribution of the remaining metrics.  
To alleviate it, we transform GM into a logarithmic GM (LGM) as follows:  
\begin{align} 
\small
\begin{split}
\label{eq:4}
LGM &= -\frac{1}{3} \Big\{ \ln\big(1-\textit{Acc}+\epsilon\big) + \ln\big(1-\text{m\_tIoU}+\epsilon\big) \\
&\quad + \ln\big(1-\text{m\_vIoU}+\epsilon\big) \Big\},
\end{split}
\end{align}
where $\epsilon$ is a small constant to prevent $\ln(0)$ when any metric reaches 1. Eq.\ref{eq:4} maps the metric range from 0 to positive infinity and ensures higher performance corresponds to a higher LGM score.
Since the logarithm transformation results in values that are typically small in magnitude, we multiply 
LGM by a linear scaling factor of 100 to ensure numerical clarity, allowing finer distinctions between different methods while preserving relative ranking.

Moreover, when the same questions appear in different CoT chains, the order in which they occur can lead to significant variations in the results. To assess the overall performance of the model across different chains, we propose the mean AM (mAM) and  mean LGM (mLGM) as follows:
\begin{align} 
\small
\label{eq:5}
mAM = \frac{1}{n}\sum_{k=1}^{n} AM_k,\quad mLGM = \frac{1}{n}\sum_{k=1}^{n} LGM_k.
\end{align}
where $n$ denotes the number of different chains. The mAM and mLGM effectively evaluate the combined impact of the various chains on the model's performance.

\vspace{-0.1cm}
\subsection{Dataset Statistics}
Here, we present detailed statistics of our dataset, including video information, meta information, qualitative analyses, and comparisons with previous works. 

\noindent\textbf{Video Information.} 
Our dataset comprises 2094 videos totalling 64.12 hours of footage. As shown in Fig.\ref{fig:statistics}(a), to ensure the inclusion of varied video genres, we categorized the videos into 9 domains: Entertainment, Daily Life, Indoor, Sports, Animals, Vehicles, Nature, Shows, and Tutorial. The length distribution of the videos, illustrated in Fig.\ref{fig:statistics}(b), demonstrates considerable diversity. The videos range in length from 15.02 seconds to 59.2 minutes with average 110.23 seconds, satisfying the requirement for diverse video lengths and better reflecting real-world scenarios. 

\noindent\textbf{Meta Information.} 
To further assess the completeness of our dataset, we assessed the meta-information annotations.  Each video is accompanied by temporal moment annotations, with an average duration of 9.06 seconds and individual durations ranging from 1.7 seconds to 47 seconds. These temporal moments account for an average of 19.3\% of the total video duration, ensuring a reasonable level of difficulty for the temporal grounding subtask. For the spatial grounding subtask, we annotated 342 objects with a total of 16,793 bounding boxes, covering approximately 19.8\% of the video resolution. This proportion is similar to that of the temporal grounding, ensuring consistent challenge levels across both tasks. Additionally, we visualized the object categories with a word cloud (Fig.\ref{fig:statistics}(c)), demonstrating that our questions robustly capture a wide diversity of objects. Tab.\ref{tab:category_comparison} provides further detailed statistics.

\begin{table*}[t]
\centering
\vspace{-0.7cm}
\setlength{\tabcolsep}{1.5pt}
\scalebox{0.85}{
\begin{tabular}{l|c|c|cc|cccc|cccc|cc}
\hline
\multirow{2}{*}{Model} & \multirow{2}{*}{Venue} & \multirow{2}{*}{Parameters} & \multicolumn{2}{c|}{What (VQA)} & \multicolumn{4}{c|}{When (Temporal Grounding)} & \multicolumn{4}{c|}{Where (Spatial Grounding)} & \multirow{2}{*}{LGM} & \multirow{2}{*}{AM} \\ 
\cline{4-13}
& & & Score & Acc & R1@0.3 & R1@0.5 & R1@0.7 & $m\_\text{tIoU}$ & AP@0.1 & AP@0.3 & AP@0.5 & $m\_\text{vIoU}$ & & \\ 
\hline
GPT-4o~\cite{OpenAIGPT4o} & - & - & \underline{1.71} & \textbf{60.78} & 23.14 & 10.35 & 5.10 & 16.67 & 19.92 & 8.36 & 2.75 & 6.47 & \textbf{39.51} & \textbf{27.97} \\ 
Gemini-2-Flash~\cite{Gemini2} & - & - & 1.59 & 53.01 & \underline{31.63} & 15.84 & \textbf{9.45} & \textbf{24.54} & 15.67 & 3.82 & 0.93 & 4.63 & \underline{36.14} & \underline{27.39} \\ 
\hline
Video-LLaMA3~\cite{zhang2025videollama} & ArXiv'25 & 7B & 1.38 & 41.94 & \textbf{35.73} & \textbf{19.80} & 8.68 & \underline{22.97} & 3.17 & 0.76 & 0.11 & 0.89 & 27.12 & 21.93 \\ 
Qwen2.5-VL~\cite{bai2025qwen2} & ArXiv'25 & 7B & 1.61 & 54.53 & 17.03 & 8.92 & 3.72 & 11.48 & \underline{35.89} & \underline{19.92} & \underline{8.36} & \underline{13.59} & 35.20 & 26.53 \\ 
Qwen2-VL~\cite{wang2024qwen2} & ArXiv'24 & 7B & 1.03 & 25.91 & 27.96 & \underline{17.94} & \underline{9.16} & 19.18 & 28.59 & 12.21 & 3.89 & 9.31 & 20.35 & 18.13 \\ 
InternVL-2.5~\cite{chen2024expanding} & ArXiv'24 & 8B & 1.46 & 44.18 & 11.98 & 4.87 & 2.34 & 8.72 & 2.18 & 0.27 & 0.04 & 0.65 & 22.69 & 17.85 \\ 
Llava-Video~\cite{zhang2024video} & ArXiv'24 & 7B & 1.50 & 49.48 & 15.12 & 6.30 & 1.43 & 10.52 & 5.23 & 0.94 & 0.18 & 1.92 & 27.11 & 20.64 \\ 
VideoChat2~\cite{li2024mvbench} & CVPR'24 & 7B & 1.27 & 36.21 & 20.47 & 13.07 & 6.49 & 13.69 & 10.06 & 1.31 & 0.14 & 2.51 & 20.74 & 17.47 \\ 
Oryx-1.5~\cite{liu2024oryx} & ICLR'25 & 7B & 0.94 & 20.47 & 17.03 & 4.48 & 1.72 & 13.54 & 35.58 & 11.60 & 2.17 & 10.14 & 16.05 & 14.72 \\ 
Video-CCAM-v1.2~\cite{fei2024video} & ArXiv'24 & 7B & \textbf{1.75} & 59.35 & 1.15 & 0.00 & 0.00 & 1.50 & - & - & - & - & 30.51 & 20.28 \\ 
\hline
TimeChat~\cite{ren2024timechat} & CVPR'24 & 7B & 1.06 & 26.38 & 17.80 & 8.68 & 3.48 & 12.01 & - & - & - & - & 14.47 & 12.80 \\ 
VTimeLLM~\cite{huang2024vtimellm} & CVPR'24 & 7B & 1.45 & 41.46 & 25.24 & 10.88 & 3.15 & 17.13 & 0.62 & 0.14 & 0.03 & 0.21 & 24.18 & 19.60 \\ 
TRACE~\cite{guo2024trace} & ICLR'25 & 7B & 0.90 & 17.60 & 28.53 & 14.17 & 6.73 & 19.74 & - & - & - & - & 13.78 & 12.45 \\ 
\hline
Sa2VA~\cite{yuan2025sa2va} & ArXiv'25 & 8B & 0.70 & 16.36 & 0.10 & 0.00 & 0.00 & 0.11 & \textbf{52.16} & \textbf{42.68} & \textbf{34.18} & \textbf{32.31} & 19.00 & 16.26 \\ 
\hline
\end{tabular}}
\vspace{-8pt}
\caption{Performance on the chain of \textit{``what-when-where"}. The top result is highlighted in \textbf{bold}, while the second is \underline{underlined}. ``-" denotes a model failed to generate formatted answers. The score ranges from 0 to 4.}
\vspace{-10pt}
\label{tab:what_when_where}
\end{table*}

\begin{table*}[t]
\centering
\setlength{\tabcolsep}{1.5pt}
\scalebox{0.85}{
\begin{tabular}{l|c|c|cc|cccc|cccc|cc}
\hline
\multirow{2}{*}{Model} & \multirow{2}{*}{Venue} & \multirow{2}{*}{Parameters} & \multicolumn{2}{c|}{What (VQA)} & \multicolumn{4}{c|}{Where (Spatial Grounding)} & \multicolumn{4}{c|}{When (Temporal Grounding)} & \multirow{2}{*}{LGM} & \multirow{2}{*}{AM} \\ 
\cline{4-13} 
& & & Score & Acc & AP@0.1 & AP@0.3 & AP@0.5 & $m\_\text{vIoU}$ & R1@0.3 & R1@0.5 & R1@0.7 & $m\_\text{tIoU}$ & & \\ 
\hline
GPT-4o~\cite{OpenAIGPT4o} & - & - & \underline{1.71} & \textbf{60.78} & 9.29 & 4.18 & 1.19 & 3.01 & 17.13 & 10.04 & 7.25 & 12.82 & \textbf{36.79} & \underline{25.53} \\ 
Gemini-2-Flash~\cite{Gemini2} & - & - & 1.59 & 53.01 & 7.49 & 1.89 & 0.58 & 2.21 & \underline{31.58} & 15.22 & \underline{8.54} & \textbf{23.83} & \underline{34.99} & \textbf{26.35} \\ 
\hline
Video-LLaMA3~\cite{zhang2025videollama} & ArXiv'25 & 7B & 1.38 & 41.94 & 0.62 & 0.17 & 0.02 & 0.19 & \textbf{35.11} & \textbf{20.42} & \textbf{9.21} & \underline{23.14} & 26.96 & 21.76 \\ 
Qwen2.5-VL~\cite{bai2025qwen2} & ArXiv'25 & 7B & 1.61 & 54.53 & 5.15 & 2.87 & \underline{1.40} & 2.00 & 11.02 & 5.39 & 2.48 & 7.61 & 29.58 & 21.38 \\ 
Qwen2-VL~\cite{wang2024qwen2} & ArXiv'24 & 7B & 1.03 & 25.91 & 7.11 & 3.55 & 1.14 & 2.41 & 24.62 & \underline{16.32} & 8.25 & 17.52 & 17.23 & 15.28 \\ 
InternVL-2.5~\cite{chen2024expanding} & ArXiv'24 & 8B & 1.46 & 44.18 & 0.42 & 0.03 & 0.00 & 0.14 & 10.83 & 3.77 & 1.57 & 7.75 & 27.15 & 17.36 \\ 
Llava-Video~\cite{zhang2024video} & ArXiv'24 & 7B & 1.50 & 49.48 & 4.29 & 1.23 & 0.25 & 1.31 & 16.89 & 5.49 & 2.00 & 12.21 & 27.54 & 21.00 \\ 
VideoChat2~\cite{li2024mvbench} & CVPR'24 & 7B & 1.27 & 36.21 & 3.08 & 0.91 & 0.30 & 0.97 & 18.08 & 12.07 & 6.20 & 12.50 & 19.77 & 16.56 \\ 
Oryx-1.5~\cite{liu2024oryx} & ICLR'25 & 7B & 0.94 & 20.47 & \underline{11.50} & \underline{4.32} & 0.96 & \underline{3.50} & 18.99 & 5.58 & 2.72 & 14.81 & 14.16 & 12.93 \\ 
Video-CCAM-v1.2~\cite{fei2024video} & ArXiv'24 & 7B & \textbf{1.75} & 59.35 & - & - & - & - & 2.19 & 0.00 & 0.00 & 2.26 & 30.88 & 20.54 \\ 
\hline
TimeChat~\cite{ren2024timechat} & CVPR'24 & 7B & 1.06 & 26.38 & - & - & - & - & 20.42 & 8.54 & 2.53 & 13.60 & 15.08 & 13.33 \\ 
VTimeLLM~\cite{huang2024vtimellm} & CVPR'24 & 7B & 1.45 & 41.46 & 0.00 & 0.00 & 0.00 & 0.00 & 8.44 & 4.53 & 2.10 & 5.96 & 19.90 & 15.81 \\ 
TRACE~\cite{guo2024trace} & ICLR'25 & 7B & 0.90 & 17.60 & - & - & - & - & 24.52 & 12.02 & 5.73 & 17.11 & 12.71 & 11.57 \\ 
\hline
Sa2VA~\cite{yuan2025sa2va} & ArXiv'25 & 8B & 0.70 & 16.36 & \textbf{58.47} & \textbf{49.47} & \textbf{40.42} & \textbf{37.48} & 0.00 & 0.00 & 0.00 & 0.00 & 21.61 & 17.95 \\ 
\hline
\end{tabular}}
\vspace{-8pt}
\caption{Performance on chain of \textit{``what-where-when"}. The top result is highlighted in \textbf{bold}, while the second is \underline{underlined}. ``-" denotes a model failed to generate formatted answers. The score ranges from 0 to 4.}
\vspace{-13pt}
\label{tab:what_where_when}
\end{table*}

\noindent\textbf{Qualitative analyses.} Fig.~\ref{fig:sample} shows an example from our V-STaR benchmark. It contains one spatio-temporal CoT reasoning thinking chain and two RSTR task chains. For each RSTR task chain, the CoT evaluation starts with a coarse-grained question about \textit{``what"} in the video. In the \textit{``what-when-where"} chain, the subsequent \textit{``when"} question incorporates the answer of `\textit{`what"} and its answer is included in the \textit{``where"} question. In the other chain, the subsequent \textit{``where"} question contains the answer of \textit{``what"} and the bounding boxes answer of \textit{``where"} will be provided without time information in the \textit{``when"} question.

\noindent\textbf{Comparisons with previous benchmarks.} We compared our V-STaR to previous Video-LLM benchmarks in Tab.~\ref{tab:benchmark-comparison}. Most existing datasets only focused on \textit{``what"} question in VQA~\cite{li2024mvbench,fang2025mmbench,fu2024video,wu2025longvideobench,chandrasegaran2025hourvideo,song2024moviechat,liu2024tempcompass}, failed to validate the model's spatio-temporal reasoning ability. Some partially cover on \textit{``when"}~\cite{lei2018tvqa,barmann2022did,xiao2024can,chen2025rextime,liu2024bench} or \textit{``where"}~\cite{munasinghe2024videoglamm}, without complete spatio-temporal reasoning chain. Only TVQA+~\cite{lei2020tvqa+} covered all of the three, but it ignored their inner spatio-temporal reasoning relationship. Instead, our V-STaR provides two CoT question chains for each sample to reveal the spatio-temporal reasoning ability of Video-LLMs.

\section{Experiments}
\subsection{Setting and Metrics} 
\noindent\textbf{Implementation Details.} We tested 14 Video-LLMs, involving 2 commercial models GPT-4o~\cite{OpenAIGPT4o} and Gemini-2-Flash~\cite{Gemini2}, and 12 open-source models. The open-source models include (i) 8 generic models: Video-LLaMA3~\cite{zhang2025videollama}, Qwen2.5-VL~\cite{bai2025qwen2}, Qwen2-VL~\cite{wang2024qwen2}, InternVL-2.5~\cite{chen2024expanding}, LLaVA-Video~\cite{zhang2024video}, VideoChat2~\cite{li2024mvbench}, Oryx-1.5~\cite{liu2024oryx}, and Video-CCAM-v1.2~\cite{fei2024video}); (ii) 3 time-aware models: TimeChat~\cite{ren2024timechat}, VTimeLLM~\cite{huang2024vtimellm}, and Trace~\cite{guo2024trace}; and (iii) 1 segmentation model, Sa2VA~\cite{yuan2025sa2va}. We followed their official configurations and sampled the video frames at 1fps for all models. If a video exceeded the model's input limitations, we applied uniform sampling to select the maximum allowable number of frames. We investigated the models' spatio-temporal reasoning ability using two RSTR task chains: \textit{``what-when-where"} and \textit{``what-where-when"}. Experiments were run on 2 NVIDIA A100 80G GPUs.


\begin{table}[!t]
    \centering
    \renewcommand{\arraystretch}{1} 
    \setlength{\tabcolsep}{1.5pt}
    \scalebox{0.70}{
    \begin{tabular}{l|cc|cc|cc|cc}
        \hline
       \multirow{2}{*}{Model} & \multicolumn{2}{c|}{Short} & \multicolumn{2}{c|}{Medium} & \multicolumn{2}{c|}{Long} & \multicolumn{2}{c}{All} \\
        \cline{2-9} 
        & mAM & mLGM & mAM & mLGM & mAM & mLGM & mAM & mLGM  \\
        \hline
        GPT-4o~\cite{OpenAIGPT4o} & \textbf{27.49} & \textbf{38.56}  & \underline{26.96} & \textbf{40.58} & 14.86 & 19.28 & \underline{26.75} & \textbf{38.15} \\
        Gemini-2-Flash~\cite{Gemini2} & 24.97 & 32.07 & \textbf{28.99} & \underline{40.35} & \textbf{37.81} & \textbf{56.14} & \textbf{26.87} & \underline{35.57} \\
        \hline
        Video-LLaMA3~\cite{zhang2025videollama} & 21.68 & 26.62 & 21.84 & 27.23 & \underline{22.46} & \underline{28.83} & 21.66 & 27.04\\
        Qwen2.5-VL~\cite{bai2025qwen2} & \underline{25.51} & \underline{34.84} & 23.67 & 32.87 & 2.20 & 2.27 & 23.96 & 32.39 \\
        Qwen2-VL~\cite{wang2024qwen2} & 15.78 & 17.50 & 18.47 & 21.22 & 14.09 & 17.53 & 16.71 & 18.79\\
        InternVL-2.5~\cite{chen2024expanding} & 17.94 & 22.90 & 17.94 & 23.06 & 9.58 & 11.19 & 17.60 & 24.92 \\
        Llava-Video~\cite{zhang2024video} & 22.37 & 30.23 & 18.28 & 22.77 & 18.23 & 25.23 & 20.82 & 27.33 \\
        VideoChat2~\cite{li2024mvbench} & 17.57 & 21.02 & 17.20 & 20.50 & 5.28 & 5.64 & 17.02 & 20.26 \\
        Oryx-1.5~\cite{liu2024oryx} & 13.17 & 14.25 & 14.83 & 16.46 & 11.89 & 13.99 & 15.11 & 13.83 \\
        Video-CCAM-v1.2~\cite{fei2024video} & 21.66 & 34.09 & 19.62 & 28.36 & 12.61 & 15.80 & 20.41 & 30.70 \\
        TimeChat~\cite{ren2024timechat} & 13.70 & 15.56 & 13.22 & 15.06 & 3.24 & 3.37 & 13.07 & 14.78 \\
        VTimeLLM~\cite{huang2024vtimellm} & 18.31 & 23.19 & 18.15 & 22.44 & 5.52 & 5.89 & 17.71 & 22.04 \\
        TRACE~\cite{guo2024trace} & 11.77 & 12.96 & 12.49 & 13.87 & 13.59 & 15.30 & 12.01 & 13.25\\
        Sa2VA~\cite{yuan2025sa2va} & 18.14 & 22.01 & 16.32 & 18.92 &8.85 & 9.70 & 17.11 & 20.31 \\
        \hline
    \end{tabular}}
    \vspace{-8pt}
    \caption{Performance on different video lengths. ``Short", ``Medium" and ``Long" denote video durations of $[0, 1]$ min, $(1, 3]$ min, and $(3, 60]$ min, respectively. The top result is highlighted in \textbf{bold}, while the second is \underline{underlined}. }
    \vspace{-15pt}
    \label{tab:comparison}
\end{table}

\begin{figure}[!t]
   \centering 
   \vspace{-3pt}
   \includegraphics[width=8.3cm]{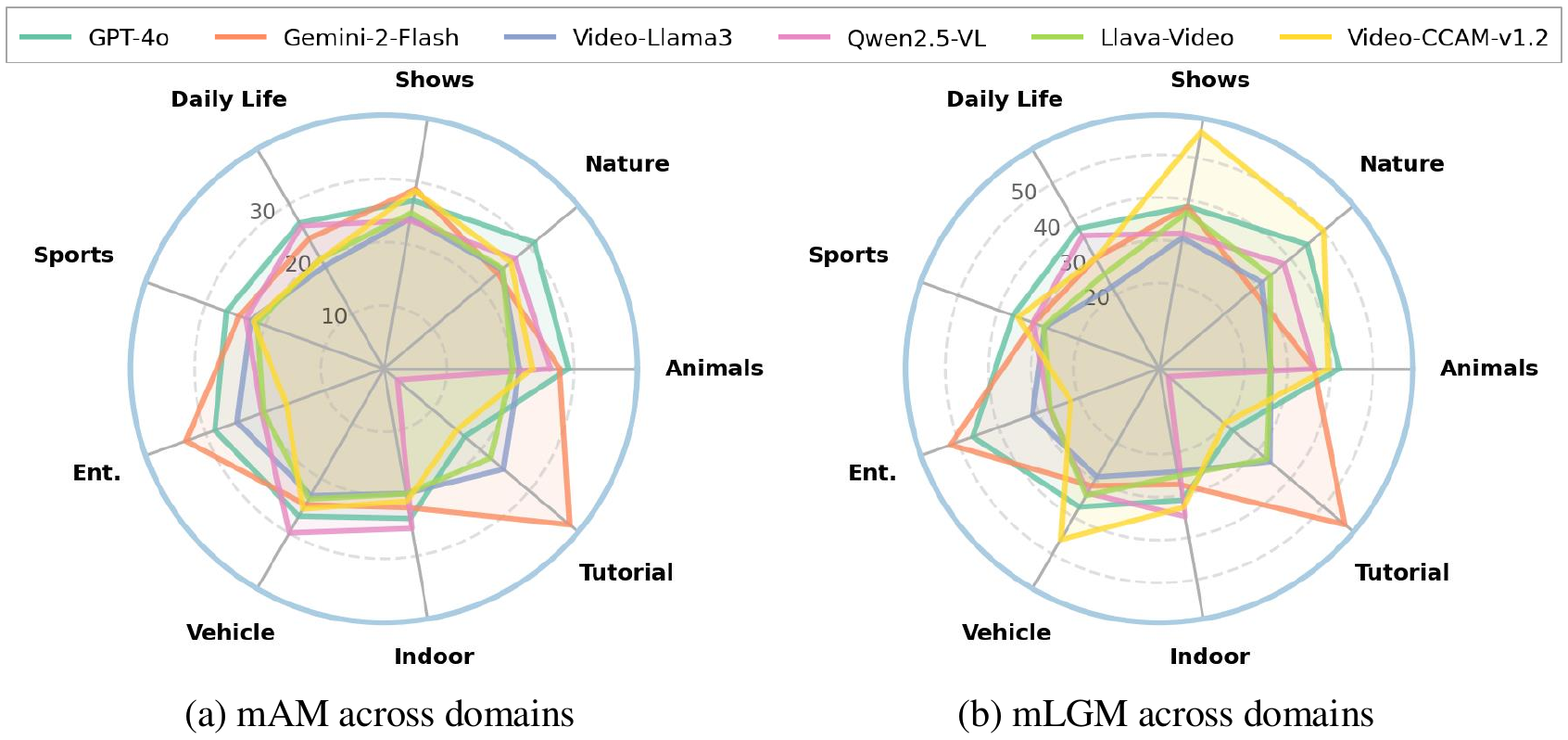}
   \vspace{-20pt}
   \caption{The performance of each domain.
   }
   \label{fig:domain}
\vspace{-20pt}
\end{figure}

\noindent\textbf{Metrics.} To evaluate the open-ended \textit{``what"} question, we follow MMBench-Video~\cite{fang2025mmbench} and use Qwen2.5-72B-Instruct to score answers from 0 to 4, denoting {\em``entirely incorrect", ``largely incorrect", ``largely correct"}, and
{\em ``entirely correct"}. Answers scoring above 2 are considered correct, allowing us to compute accuracy. For the \textit{``when"} question, we follow the commonly used temporal grounding metrics, “$\text{R}@\text{n}, \text{tIoU=}m$", which refers to the percentage of top-$n$ prediction with temporal IoU score larger than $m$, and mean temporal IoU score ($\text{m\_tIoU}$). For the \textit{``where"} question, we follow TVQA+~\cite{lei2020tvqa+} and VidSTG~\cite{zhang2020does} to use the Average Precision score ($\text{AP@vIoU=}m$) and mean visual Intersection over Union ($\text{m\_vIoU}$) of every annotated frame. We follow the proposed LGM (Eq.\ref{eq:4}) and AM (Eq.\ref{eq:1}) to measure a model's spatial-temporal reasoning ability. A higher LGM indicates a better overall spatio-temporal reasoning ability of the model, and a higher AM indicates a more average performance of the model on the three metrics.

\subsection{Quantitative Results}

\noindent\textbf{Performance on \textit{``what-when-where"} chain.} As shown in Tab.\ref{tab:what_when_where}, the \textit{``what-where-when"} chain evaluates a model's spatial-temporal reasoning ability. Overall, GPT-4o, Gemini-2-Flash, and Qwen2.5-VL demonstrate the strongest spatio-temporal reasoning capabilities, ranking as the top-3 models. Their scores for LGM and AM are 39.15/36.14/35.20 and 27.97\%/27.39\%/26.53\%, respectively. At the lower end, Trace, TimeChat, and Oryx-1.5 rank as the bottom-3 models, with LGM and AM scores of 13.78/14.47/16.05 and 12.45\%/12.80\%/14.72\%, respectively. The remaining models, ranked in descending order based on their LGM scores, are Video-CCAM, Video-LLaMA3, LLaVA-Video, VTimeLLM, InternVL-2.5, VideoChat2, Qwen2-VL, and SA2VA. Among them, Video-LLaMA3 demonstrates the most balanced performance, with an AM score of 21.93\%.

In open-source models, Video-CCAM-v1.2 leads in VQA accuracy (59.35\%) but struggles with fine-grained temporal ($\text{m\_tIoU}$:1.50\%) and spatial understanding (fail). While VideoLLaMA3 leads in temporal grounding ($\text{m\_tIoU}$:22.97\%), it lacks consistency across the other two tasks. Sa2VA excels in spatial grounding ($\text{m\_vIoU}$: 32.31\%), but performs poorly in VQA (16.36\%) and temporal grounding ($\text{m\_tIoU}$: 0.11\%). In contrast, Qwen2.5-VL shows the most balanced performance across all three tasks, leading open-source models in overall performance. They highlight that maintaining consistency across \textit{``what-when-where"} reasoning is crucial, as weaknesses in earlier steps propagate and affect overall performance.

%


\noindent\textbf{Performance on \textit{``what-where-when"} chain.} Tab.\ref{tab:what_where_when} presents the models' performance across the other reasoning chain \textit{``what-where-when"}. In this chain, GPT-4o and Gemini-2-Flash achieve the top-2 overall performances, with LGM scores of 36.79 and 34.99, respectively. However, Gemini-2-Flash exhibits a more balanced performance than GPT-4o, with an AM score of 26.35\% compared to 25.53\%. Although Video-CCAM-v1.2 outperforms Qwen2.5-VL in overall performance (LGM: 30.88 vs. 29.58), it is less consistent across tasks (AM: 20.54\% vs. 21.38\%). The bottom-3 models remain Trace, Oryx-1.5, and TimeChat, with LGM and AM scores of (12.71/14.16/15.08) and (11.57\%/12.93\%/13.33\%), respectively. The remaining models, ranked by LGM from high to low, are LLaVA-Video, InternVL-2.5, Video-LLaMA3, SA2VA, VTimeLLM, VideoChat-2, and Qwen2-VL.

In this reasoning chain, without temporal grounding as a prerequisite step, models exhibit a general performance drop in spatial grounding. The most significant decline is observed in Qwen2.5-VL, whose $\text{m\_vIoU}$ score drops sharply to 2.00\%. In temporal grounding, excessive spatial information in the prompts leads to a substantial performance drop in VTimeLLM, reducing its $\text{m\_tIoU}$ score to 5.96\%. Interestingly, Llava-video, Video-LLaMA3, Oryx-1.5 and TimeChat show slight improvements in this setting.

\begin{table}[!t]
\renewcommand{\arraystretch}{1}
\setlength{\tabcolsep}{1pt}
\vspace{-3pt}
\centering
\scalebox{0.63}{
\begin{tabular}{l|ccc|ccc|ccc}
\hline
\multirow{2}{*}{Model} 
& \multicolumn{3}{c|}{Acc\textbar tIoU@0.3} & \multicolumn{3}{c|}{Acc\textbar vIoU@0.1}  & \multicolumn{3}{c}{Acc\textbar tIoU@0.3, vIoU@0.1} \\ 
\cline{2-10} 
 & Chain 1 & Chain 2 & $|\Delta|$ & Chain 1 & Chain 2 & $|\Delta|$ & Chain 1 & Chain 2 & $|\Delta|$  \\ 
\hline
GPT-4o~\cite{OpenAIGPT4o}     & 15.12 & 11.16 & \textbf{3.96}  & 15.27 & \textbf{7.59}  & 7.68  & 4.53  & \textbf{3.91}  & 0.62  \\
Gemini-2-Flash~\cite{Gemini2}  & \textbf{19.70}  & \textbf{19.04} & 0.66 & 8.68   & 4.48 & 4.2  & 3.48   & 2.24 & 1.24  \\ 
\hline
Video-LLaMA3~\cite{zhang2025videollama} & 15.41 & 14.89 & 0.52 & 1.34  & 0.19 & 1.15  & 0.52   & 0.05 & 0.47  \\
Qwen2.5-VL~\cite{bai2025qwen2} & 10.73  & 7.20 & 3.53  & \textbf{24.24}  & 4.25 & \textbf{19.99}  & \textbf{4.68} & 1.15 & \textbf{3.53}  \\
Qwen2-VL~\cite{wang2024qwen2}  & 8.06   & 6.68 & 1.38  & 7.11  & 2.05  & 5.06 & 2.29   & 1.53 & 0.76  \\
InternVL-2.5~\cite{chen2024expanding} & 5.92   & 4.77  & 1.15  & 0.81   & 0.19 & 0.62 & 0.19 & 0.05 & 0.14 \\
Llava-Video~\cite{zhang2024video} & 7.92   & 8.97 & 1.05  & 3.05 & 2.86 & 0.19 & 0.67   & 0.86  & 0.19 \\
VideoChat-2~\cite{li2024mvbench}  & 8.78   & 7.73 & 1.05  & 3.34   & 1.24 & 2.1 & 0.29   & 0.62 & 0.33  \\
Oryx-1.5~\cite{liu2024oryx} & 3.58   & 3.77  & 0.19  & 6.25   & 2.05 & 4.2 & 1.24 & 0.67 & 0.57\\
\hline
\end{tabular}}
\vspace{-8pt}
\caption{Joint performance evaluation across models.}
\label{tab:joint}
\vspace{-22pt}
\end{table}

\begin{figure*}[!t]
   \centering 
   \vspace{-20pt}
   \includegraphics[width=0.94\textwidth]{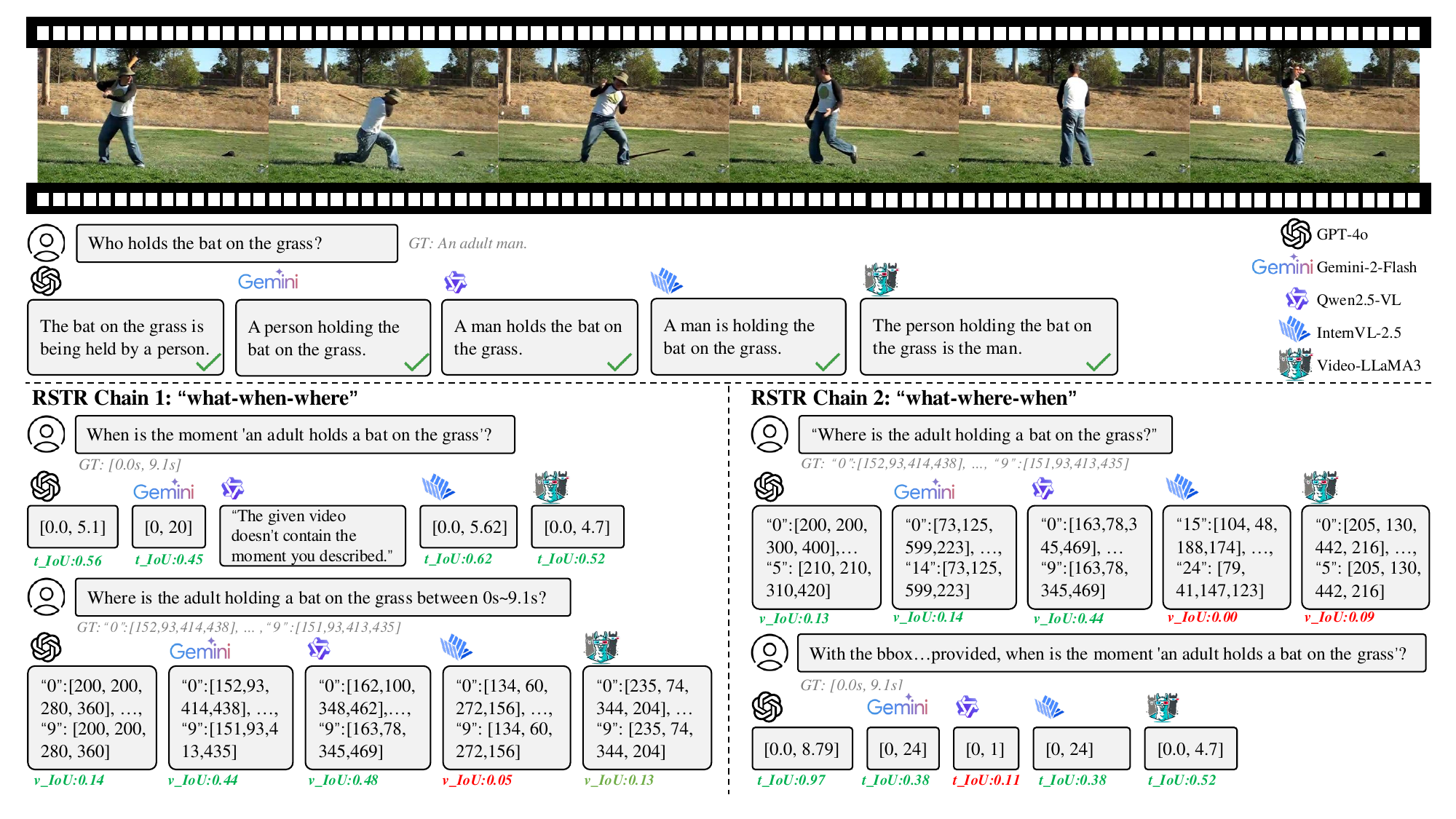}
   \vspace{-10pt}
   \caption{An example showcasing the performance of five models. 
   }
   \label{fig:vis}
\vspace{-10pt}
\end{figure*}
\noindent\textbf{Performance on each domain.} We visualize the performance on each domain in Fig.\ref{fig:domain} using mAM (left) and mLGM (right). 
Gemini-2-Flash (orange) and LLaVA-Video (green) demonstrate relatively balanced performance across domains. GPT-4o (blue-green) performs best in the Animals, Nature, Daily Life, and Sports but lags in the Tutorial. Qwen2.5-VL (pink) shows strong performance in Vehicle but also lags in Tutorials, whereas Video-CCAM-v1.2 (yellow) shows a strong advantage in Shows, Vehicles, and Nature domains but weaker performance in others. Overall, it indicates that current Video-LLMs do not generalize well across all domains, emphasizing the need for domain-specific evaluation in spatio-temporal reasoning tasks.

\noindent\textbf{Effect of different video length.} From the results in Table \ref{tab:comparison}, the models' spatial-temporal reasoning ability is evaluated across different video lengths, showing how performance shifts as reasoning complexity increases.
GPT-4o performs well on short videos (mLGM: 38.56, mAM: 27.49\%) but struggles with long sequences, suggesting weaker long-range dependency modelling. In medium videos, GPT-4o achieves the highest performance with 40.58\% in mLGM, while Gemini-2-Flash demonstrates greater balance with 28.99\% in mAM. Gemini-2-Flash consistently outperforms others in long videos, achieving the highest mLGM (56.14) and mAM (37.81\%), indicating strong temporal reasoning over extended sequences. It also leads in medium (mLGM: 40.35, mAM: 28.99\%) and overall performance (mLGM: 35.57, mAM: 26.87\%), highlighting its robustness across different durations. Overall, GPT-4o achieves the highest overall performance with 38.15 in mLGM, with a competitive mAM score.
Among open-source models, Qwen2.5-VL, InternVL-2.5 and Video-Chat2 perform moderately well but show noticeable declines as video length increases. Video-CCAM-v1.2, TimeChat, and TRACE struggle across all durations, with mLGM scores below 15, indicating weak spatial-temporal integration. These results suggest that handling longer sequences remains a challenge, requiring models to improve long-range dependency modelling to maintain reasoning continuity across extended video durations.

\noindent\textbf{Results on joint performance.} 
To further evaluate spatio-temporal reasoning, we analyze models' joint performance on RSTR tasks in Tab.\ref{tab:joint}. Using thresholds ($\text{tIoU}=0.3$, $\text{vIoU}=0.1$), we measure the percentage of samples where models correctly use temporal, spatial, or both cues to infer answers. Results show that Gemini-2-Flash excels in temporal reasoning, while Qwen2.5-VL leads in spatial reasoning for Chain 1 and SA2VA for Chain 2. For combined spatio-temporal reasoning, Qwen2.5-VL ranks highest in Chain 1, and Gemini-2 in Chain 2, but both still with low accuracy (4.68\% and 2.24\%). This highlights the limited spatio-temporal reasoning abilities of current Video-LLMs. From the changes between the two chains in the joint performance, GPT-4o and Qwen2.5-VL are the most affected.

\noindent\textbf{Qualitative analysis.} Fig.\ref{fig:vis} presents a visualization of five models’ performance on the V-STaR benchmark. While all correctly answer the \textit{``what"} questions, their spatio-temporal reasoning remains weak. In the \textit{``what-when-where"} chain, Qwen2.5-VL achieves the best spatial grounding but struggles with temporal localization, whereas InternVL-2.5 excels in temporal grounding but fails in spatial accuracy. GPT-4o, Gemini-2-Flash, and VideoLlama3 show a more balanced understanding of both aspects. In the \textit{``what-where-when"} chain, Qwen2.5-VL maintains stable spatial grounding despite missing temporal cues, while others degrade. When given spatial information, GPT-4o and Qwen2.5-VL improve in temporal grounding, VideoLlama3 remains unchanged, and Gemini-2-Flash and InternVL-2.5 perform worse. Notably, Video-LLMs often analyse each frame independently, overlooking dynamic relationships among frames and treating objects as static, revealing a key limitation in their motion perception. 

\noindent\textbf{Supplementary material.}
 App. A and B provides more codes and benchmark details. App. C includes additional implementation information. App. D presents an in-depth experiment results with 24 tables, and App. E is limitation. 
 \vspace{-20pt}
\section{Conclusion}
 \vspace{-5pt}
This work introduces a new Video-LLM spatio-temporal reasoning benchmark, V-STaR, the first benchmark for comprehensively assessing spatio-temporal reasoning ability of Video-LLMs. We constructed a dataset with coarse-to-fine CoT questions for structured evaluation and introduced a new Logarithmic Geometric Mean (LGM) metric for scoring video spatio-temporal reasoning performance. Experiments on 14 Video-LLMs provide insights into their reasoning capabilities and future improvements.

{
    \small
    \bibliographystyle{ieeenat_fullname}
    \bibliography{main}

\begin{thebibliography}{49}
\providecommand{\natexlab}[1]{#1}
\providecommand{\url}[1]{\texttt{#1}}
\expandafter\ifx\csname urlstyle\endcsname\relax
  \providecommand{\doi}[1]{doi: #1}\else
  \providecommand{\doi}{doi: \begingroup \urlstyle{rm}\Url}\fi

\bibitem[Bai et~al.(2025)Bai, Chen, Liu, Wang, Ge, Song, Dang, Wang, Wang, Tang, et~al.]{bai2025qwen2}
Shuai Bai, Keqin Chen, Xuejing Liu, Jialin Wang, Wenbin Ge, Sibo Song, Kai Dang, Peng Wang, Shijie Wang, Jun Tang, et~al.
\newblock Qwen2. 5-vl technical report.
\newblock \emph{arXiv preprint arXiv:2502.13923}, 2025.

\bibitem[B{\"a}rmann and Waibel(2022)]{barmann2022did}
Leonard B{\"a}rmann and Alex Waibel.
\newblock Where did i leave my keys?-episodic-memory-based question answering on egocentric videos.
\newblock In \emph{Proceedings of the IEEE/CVF Conference on Computer Vision and Pattern Recognition}, pages 1560--1568, 2022.

\bibitem[Chandrasegaran et~al.(2025)Chandrasegaran, Gupta, Hadzic, Kota, He, Eyzaguirre, Durante, Li, Wu, and Li]{chandrasegaran2025hourvideo}
Keshigeyan Chandrasegaran, Agrim Gupta, Lea~M Hadzic, Taran Kota, Jimming He, Crist{\'o}bal Eyzaguirre, Zane Durante, Manling Li, Jiajun Wu, and Fei-Fei Li.
\newblock Hourvideo: 1-hour video-language understanding.
\newblock \emph{Advances in Neural Information Processing Systems}, 37:\penalty0 53168--53197, 2025.

\bibitem[Chen et~al.(2025)Chen, Liao, Lin, Yu, Chen, and Wang]{chen2025rextime}
Jr-Jen Chen, Yu-Chien Liao, Hsi-Che Lin, Yu-Chu Yu, Yen-Chun Chen, and Frank Wang.
\newblock Rextime: A benchmark suite for reasoning-across-time in videos.
\newblock \emph{Advances in Neural Information Processing Systems}, 37:\penalty0 28662--28673, 2025.

\bibitem[Chen et~al.(2023)Chen, Sikka, Cogswell, Ji, and Divakaran]{chen2023measuring}
Yangyi Chen, Karan Sikka, Michael Cogswell, Heng Ji, and Ajay Divakaran.
\newblock Measuring and improving chain-of-thought reasoning in vision-language models.
\newblock \emph{arXiv preprint arXiv:2309.04461}, 2023.

\bibitem[Chen et~al.(2024)Chen, Wang, Cao, Liu, Gao, Cui, Zhu, Ye, Tian, Liu, et~al.]{chen2024expanding}
Zhe Chen, Weiyun Wang, Yue Cao, Yangzhou Liu, Zhangwei Gao, Erfei Cui, Jinguo Zhu, Shenglong Ye, Hao Tian, Zhaoyang Liu, et~al.
\newblock Expanding performance boundaries of open-source multimodal models with model, data, and test-time scaling.
\newblock \emph{arXiv preprint arXiv:2412.05271}, 2024.

\bibitem[Fang et~al.(2025)Fang, Mao, Duan, Zhao, Li, Lin, and Chen]{fang2025mmbench}
Xinyu Fang, Kangrui Mao, Haodong Duan, Xiangyu Zhao, Yining Li, Dahua Lin, and Kai Chen.
\newblock Mmbench-video: A long-form multi-shot benchmark for holistic video understanding.
\newblock \emph{Advances in Neural Information Processing Systems}, 37:\penalty0 89098--89124, 2025.

\bibitem[Fei et~al.(2024)Fei, Li, Deng, Wang, Liu, and Wang]{fei2024video}
Jiajun Fei, Dian Li, Zhidong Deng, Zekun Wang, Gang Liu, and Hui Wang.
\newblock Video-ccam: Enhancing video-language understanding with causal cross-attention masks for short and long videos.
\newblock \emph{arXiv preprint arXiv:2408.14023}, 2024.

\bibitem[Fu et~al.(2024)Fu, Dai, Luo, Li, Ren, Zhang, Wang, Zhou, Shen, Zhang, et~al.]{fu2024video}
Chaoyou Fu, Yuhan Dai, Yongdong Luo, Lei Li, Shuhuai Ren, Renrui Zhang, Zihan Wang, Chenyu Zhou, Yunhang Shen, Mengdan Zhang, et~al.
\newblock Video-mme: The first-ever comprehensive evaluation benchmark of multi-modal llms in video analysis.
\newblock \emph{arXiv preprint arXiv:2405.21075}, 2024.

\bibitem[Gao et~al.(2017)Gao, Sun, Yang, and Nevatia]{gao2017tall}
Jiyang Gao, Chen Sun, Zhenheng Yang, and Ram Nevatia.
\newblock Tall: Temporal activity localization via language query.
\newblock In \emph{Proceedings of the IEEE international conference on computer vision}, pages 5267--5275, 2017.

\bibitem[{Google}(2024)]{Gemini2}
{Google}.
\newblock Google, gemini-2-flash.
\newblock Technical report, Google, 2024.

\bibitem[Guo et~al.(2025)Guo, Yang, Zhang, Song, Zhang, Xu, Zhu, Ma, Wang, Bi, et~al.]{guo2025deepseek}
Daya Guo, Dejian Yang, Haowei Zhang, Junxiao Song, Ruoyu Zhang, Runxin Xu, Qihao Zhu, Shirong Ma, Peiyi Wang, Xiao Bi, et~al.
\newblock Deepseek-r1: Incentivizing reasoning capability in llms via reinforcement learning.
\newblock \emph{arXiv preprint arXiv:2501.12948}, 2025.

\bibitem[Guo et~al.(2024)Guo, Liu, Li, Tang, Liu, and Chen]{guo2024trace}
Yongxin Guo, Jingyu Liu, Mingda Li, Xiaoying Tang, Qingbin Liu, and Xi Chen.
\newblock Trace: Temporal grounding video llm via causal event modeling.
\newblock \emph{arXiv preprint arXiv:2410.05643}, 2024.

\bibitem[Hu et~al.(2025)Hu, Lin, Yan, and Gong]{hu2025leveraging}
Jian Hu, Jiayi Lin, Junchi Yan, and Shaogang Gong.
\newblock Leveraging hallucinations to reduce manual prompt dependency in promptable segmentation.
\newblock \emph{Advances in Neural Information Processing Systems}, 37:\penalty0 107171--107197, 2025.

\bibitem[Huang et~al.(2024)Huang, Wang, Chen, Song, and Zhu]{huang2024vtimellm}
Bin Huang, Xin Wang, Hong Chen, Zihan Song, and Wenwu Zhu.
\newblock Vtimellm: Empower llm to grasp video moments.
\newblock In \emph{Proceedings of the IEEE/CVF Conference on Computer Vision and Pattern Recognition}, pages 14271--14280, 2024.

\bibitem[Huang et~al.(2019)Huang, Zhao, and Huang]{huang2019got}
Lianghua Huang, Xin Zhao, and Kaiqi Huang.
\newblock Got-10k: A large high-diversity benchmark for generic object tracking in the wild.
\newblock \emph{IEEE transactions on pattern analysis and machine intelligence}, 43\penalty0 (5):\penalty0 1562--1577, 2019.

\bibitem[Krishna et~al.(2017)Krishna, Hata, Ren, Fei-Fei, and Carlos~Niebles]{krishna2017dense}
Ranjay Krishna, Kenji Hata, Frederic Ren, Li Fei-Fei, and Juan Carlos~Niebles.
\newblock Dense-captioning events in videos.
\newblock In \emph{Proceedings of the IEEE international conference on computer vision}, pages 706--715, 2017.

\bibitem[Lanham et~al.(2023)Lanham, Chen, Radhakrishnan, Steiner, Denison, Hernandez, Li, Durmus, Hubinger, Kernion, et~al.]{lanham2023measuring}
Tamera Lanham, Anna Chen, Ansh Radhakrishnan, Benoit Steiner, Carson Denison, Danny Hernandez, Dustin Li, Esin Durmus, Evan Hubinger, Jackson Kernion, et~al.
\newblock Measuring faithfulness in chain-of-thought reasoning.
\newblock \emph{arXiv preprint arXiv:2307.13702}, 2023.

\bibitem[Lei et~al.(2018)Lei, Yu, Bansal, and Berg]{lei2018tvqa}
Jie Lei, Licheng Yu, Mohit Bansal, and Tamara Berg.
\newblock Tvqa: Localized, compositional video question answering.
\newblock In \emph{Proceedings of the 2018 Conference on Empirical Methods in Natural Language Processing}, pages 1369--1379, 2018.

\bibitem[Lei et~al.(2020)Lei, Yu, Berg, and Bansal]{lei2020tvqa+}
Jie Lei, Licheng Yu, Tamara Berg, and Mohit Bansal.
\newblock Tvqa+: Spatio-temporal grounding for video question answering.
\newblock In \emph{Proceedings of the 58th Annual Meeting of the Association for Computational Linguistics}, pages 8211--8225, 2020.

\bibitem[Li et~al.(2022)Li, Niu, and Zhang]{li2022representation}
Jiangtong Li, Li Niu, and Liqing Zhang.
\newblock From representation to reasoning: Towards both evidence and commonsense reasoning for video question-answering.
\newblock In \emph{Proceedings of the IEEE/CVF conference on computer vision and pattern recognition}, pages 21273--21282, 2022.

\bibitem[Li et~al.(2023)Li, He, Wang, Li, Wang, Luo, Wang, Wang, and Qiao]{li2023videochat}
KunChang Li, Yinan He, Yi Wang, Yizhuo Li, Wenhai Wang, Ping Luo, Yali Wang, Limin Wang, and Yu Qiao.
\newblock Videochat: Chat-centric video understanding.
\newblock \emph{arXiv preprint arXiv:2305.06355}, 2023.

\bibitem[Li et~al.(2024{\natexlab{a}})Li, Wang, He, Li, Wang, Liu, Wang, Xu, Chen, Luo, et~al.]{li2024mvbench}
Kunchang Li, Yali Wang, Yinan He, Yizhuo Li, Yi Wang, Yi Liu, Zun Wang, Jilan Xu, Guo Chen, Ping Luo, et~al.
\newblock Mvbench: A comprehensive multi-modal video understanding benchmark.
\newblock In \emph{Proceedings of the IEEE/CVF Conference on Computer Vision and Pattern Recognition}, pages 22195--22206, 2024{\natexlab{a}}.

\bibitem[Li et~al.(2024{\natexlab{b}})Li, Chen, Hu, Wang, Shi, and Zhang]{li2024videovista}
Yunxin Li, Xinyu Chen, Baotian Hu, Longyue Wang, Haoyuan Shi, and Min Zhang.
\newblock Videovista: A versatile benchmark for video understanding and reasoning.
\newblock \emph{arXiv preprint arXiv:2406.11303}, 2024{\natexlab{b}}.

\bibitem[Lin et~al.(2014)Lin, Maire, Belongie, Hays, Perona, Ramanan, Doll{\'a}r, and Zitnick]{lin2014microsoft}
Tsung-Yi Lin, Michael Maire, Serge Belongie, James Hays, Pietro Perona, Deva Ramanan, Piotr Doll{\'a}r, and C~Lawrence Zitnick.
\newblock Microsoft coco: Common objects in context.
\newblock In \emph{Computer vision--ECCV 2014: 13th European conference, zurich, Switzerland, September 6-12, 2014, proceedings, part v 13}, pages 740--755. Springer, 2014.

\bibitem[Liu et~al.(2024{\natexlab{a}})Liu, Li, Liu, Wang, Ren, Li, Chen, Sun, and Hou]{liu2024tempcompass}
Yuanxin Liu, Shicheng Li, Yi Liu, Yuxiang Wang, Shuhuai Ren, Lei Li, Sishuo Chen, Xu Sun, and Lu Hou.
\newblock Tempcompass: Do video llms really understand videos?
\newblock \emph{arXiv preprint arXiv:2403.00476}, 2024{\natexlab{a}}.

\bibitem[Liu et~al.(2024{\natexlab{b}})Liu, Ma, Qi, Wu, Shan, and Chen]{liu2024bench}
Ye Liu, Zongyang Ma, Zhongang Qi, Yang Wu, Ying Shan, and Chang~Wen Chen.
\newblock Et bench: Towards open-ended event-level video-language understanding.
\newblock \emph{arXiv preprint arXiv:2409.18111}, 2024{\natexlab{b}}.

\bibitem[Liu et~al.(2024{\natexlab{c}})Liu, Dong, Liu, Hu, Lu, and Rao]{liu2024oryx}
Zuyan Liu, Yuhao Dong, Ziwei Liu, Winston Hu, Jiwen Lu, and Yongming Rao.
\newblock Oryx mllm: On-demand spatial-temporal understanding at arbitrary resolution.
\newblock \emph{arXiv preprint arXiv:2409.12961}, 2024{\natexlab{c}}.

\bibitem[Munasinghe et~al.(2023)Munasinghe, Thushara, Maaz, Rasheed, Khan, Shah, and Khan]{munasinghe2023pg}
Shehan Munasinghe, Rusiru Thushara, Muhammad Maaz, Hanoona~Abdul Rasheed, Salman Khan, Mubarak Shah, and Fahad Khan.
\newblock Pg-video-llava: Pixel grounding large video-language models.
\newblock \emph{arXiv preprint arXiv:2311.13435}, 2023.

\bibitem[Munasinghe et~al.(2024)Munasinghe, Gani, Zhu, Cao, Xing, Khan, and Khan]{munasinghe2024videoglamm}
Shehan Munasinghe, Hanan Gani, Wenqi Zhu, Jiale Cao, Eric Xing, Fahad~Shahbaz Khan, and Salman Khan.
\newblock Videoglamm: A large multimodal model for pixel-level visual grounding in videos.
\newblock \emph{arXiv preprint arXiv:2411.04923}, 2024.

\bibitem[{OpenAI}(2023)]{OpenAIGPT4}
{OpenAI}.
\newblock Gpt-4 technical report.
\newblock Technical report, OpenAI, 2023.

\bibitem[{OpenAI}(March 2024)]{OpenAIGPT4o}
{OpenAI}.
\newblock Openai, gpt-40.
\newblock Technical report, March 2024.

\bibitem[Ren et~al.(2024)Ren, Yao, Li, Sun, and Hou]{ren2024timechat}
Shuhuai Ren, Linli Yao, Shicheng Li, Xu Sun, and Lu Hou.
\newblock Timechat: A time-sensitive multimodal large language model for long video understanding.
\newblock In \emph{Proceedings of the IEEE/CVF Conference on Computer Vision and Pattern Recognition}, pages 14313--14323, 2024.

\bibitem[Signorelli et~al.(2020)Signorelli, D{\"u}ndar-Coecke, Wang, and Coecke]{signorelli2020cognitive}
Camilo~Miguel Signorelli, Selma D{\"u}ndar-Coecke, Vincent Wang, and Bob Coecke.
\newblock Cognitive structures of space-time.
\newblock \emph{Frontiers in Psychology}, 11:\penalty0 527114, 2020.

\bibitem[Song et~al.(2024)Song, Chai, Wang, Zhang, Zhou, Wu, Chi, Guo, Ye, Zhang, et~al.]{song2024moviechat}
Enxin Song, Wenhao Chai, Guanhong Wang, Yucheng Zhang, Haoyang Zhou, Feiyang Wu, Haozhe Chi, Xun Guo, Tian Ye, Yanting Zhang, et~al.
\newblock Moviechat: From dense token to sparse memory for long video understanding.
\newblock In \emph{Proceedings of the IEEE/CVF Conference on Computer Vision and Pattern Recognition}, pages 18221--18232, 2024.

\bibitem[Sprague et~al.(2024)Sprague, Yin, Rodriguez, Jiang, Wadhwa, Singhal, Zhao, Ye, Mahowald, and Durrett]{sprague2024cot}
Zayne Sprague, Fangcong Yin, Juan~Diego Rodriguez, Dongwei Jiang, Manya Wadhwa, Prasann Singhal, Xinyu Zhao, Xi Ye, Kyle Mahowald, and Greg Durrett.
\newblock To cot or not to cot? chain-of-thought helps mainly on math and symbolic reasoning.
\newblock \emph{arXiv preprint arXiv:2409.12183}, 2024.

\bibitem[Tang et~al.(2021)Tang, Liao, Liu, Li, Jin, Jiang, Yu, and Xu]{tang2021human}
Zongheng Tang, Yue Liao, Si Liu, Guanbin Li, Xiaojie Jin, Hongxu Jiang, Qian Yu, and Dong Xu.
\newblock Human-centric spatio-temporal video grounding with visual transformers.
\newblock \emph{IEEE Transactions on Circuits and Systems for Video Technology}, 32\penalty0 (12):\penalty0 8238--8249, 2021.

\bibitem[Wang et~al.(2024)Wang, Bai, Tan, Wang, Fan, Bai, Chen, Liu, Wang, Ge, et~al.]{wang2024qwen2}
Peng Wang, Shuai Bai, Sinan Tan, Shijie Wang, Zhihao Fan, Jinze Bai, Keqin Chen, Xuejing Liu, Jialin Wang, Wenbin Ge, et~al.
\newblock Qwen2-vl: Enhancing vision-language model's perception of the world at any resolution.
\newblock \emph{arXiv preprint arXiv:2409.12191}, 2024.

\bibitem[Wu et~al.(2025)Wu, Li, Chen, and Li]{wu2025longvideobench}
Haoning Wu, Dongxu Li, Bei Chen, and Junnan Li.
\newblock Longvideobench: A benchmark for long-context interleaved video-language understanding.
\newblock \emph{Advances in Neural Information Processing Systems}, 37:\penalty0 28828--28857, 2025.

\bibitem[Xiao et~al.(2024)Xiao, Yao, Li, and Chua]{xiao2024can}
Junbin Xiao, Angela Yao, Yicong Li, and Tat-Seng Chua.
\newblock Can i trust your answer? visually grounded video question answering.
\newblock In \emph{Proceedings of the IEEE/CVF Conference on Computer Vision and Pattern Recognition}, pages 13204--13214, 2024.

\bibitem[Yan et~al.(2024)Yan, Wang, Yan, Jiang, Hu, Kang, Xie, and Gavves]{yan2024visa}
Cilin Yan, Haochen Wang, Shilin Yan, Xiaolong Jiang, Yao Hu, Guoliang Kang, Weidi Xie, and Efstratios Gavves.
\newblock Visa: Reasoning video object segmentation via large language models.
\newblock In \emph{European Conference on Computer Vision}, pages 98--115. Springer, 2024.

\bibitem[Yu et~al.(2016)Yu, Poirson, Yang, Berg, and Berg]{yu2016modeling}
Licheng Yu, Patrick Poirson, Shan Yang, Alexander~C Berg, and Tamara~L Berg.
\newblock Modeling context in referring expressions.
\newblock In \emph{Computer Vision--ECCV 2016: 14th European Conference, Amsterdam, The Netherlands, October 11-14, 2016, Proceedings, Part II 14}, pages 69--85. Springer, 2016.

\bibitem[Yuan et~al.(2025)Yuan, Li, Zhang, Huang, Xu, Ji, Tong, Qi, Feng, and Yang]{yuan2025sa2va}
Haobo Yuan, Xiangtai Li, Tao Zhang, Zilong Huang, Shilin Xu, Shunping Ji, Yunhai Tong, Lu Qi, Jiashi Feng, and Ming-Hsuan Yang.
\newblock Sa2va: Marrying sam2 with llava for dense grounded understanding of images and videos.
\newblock \emph{arXiv preprint arXiv:2501.04001}, 2025.

\bibitem[Zhang et~al.(2025)Zhang, Li, Cheng, Hu, Yuan, Chen, Leng, Jiang, Zhang, Li, et~al.]{zhang2025videollama}
Boqiang Zhang, Kehan Li, Zesen Cheng, Zhiqiang Hu, Yuqian Yuan, Guanzheng Chen, Sicong Leng, Yuming Jiang, Hang Zhang, Xin Li, et~al.
\newblock Videollama 3: Frontier multimodal foundation models for image and video understanding.
\newblock \emph{arXiv preprint arXiv:2501.13106}, 2025.

\bibitem[Zhang et~al.(2020{\natexlab{a}})Zhang, Sun, Jing, and Zhou]{zhang2020span}
Hao Zhang, Aixin Sun, Wei Jing, and Joey~Tianyi Zhou.
\newblock Span-based localizing network for natural language video localization.
\newblock \emph{arXiv preprint arXiv:2004.13931}, 2020{\natexlab{a}}.

\bibitem[Zhang et~al.(2020{\natexlab{b}})Zhang, Peng, Fu, and Luo]{zhang2020learning}
Songyang Zhang, Houwen Peng, Jianlong Fu, and Jiebo Luo.
\newblock Learning 2d temporal adjacent networks for moment localization with natural language.
\newblock In \emph{Proceedings of the AAAI conference on artificial intelligence}, pages 12870--12877, 2020{\natexlab{b}}.

\bibitem[Zhang et~al.(2024)Zhang, Wu, Li, Li, Ma, Liu, and Li]{zhang2024video}
Yuanhan Zhang, Jinming Wu, Wei Li, Bo Li, Zejun Ma, Ziwei Liu, and Chunyuan Li.
\newblock Video instruction tuning with synthetic data.
\newblock \emph{arXiv preprint arXiv:2410.02713}, 2024.

\bibitem[Zhang et~al.(2020{\natexlab{c}})Zhang, Zhao, Zhao, Wang, Liu, and Gao]{zhang2020does}
Zhu Zhang, Zhou Zhao, Yang Zhao, Qi Wang, Huasheng Liu, and Lianli Gao.
\newblock Where does it exist: Spatio-temporal video grounding for multi-form sentences.
\newblock In \emph{Proceedings of the IEEE/CVF Conference on Computer Vision and Pattern Recognition}, pages 10668--10677, 2020{\natexlab{c}}.

\bibitem[Zhou et~al.(2024)Zhou, Shu, Zhao, Wu, Xiao, Yang, Xiong, Zhang, Huang, and Liu]{zhou2024mlvu}
Junjie Zhou, Yan Shu, Bo Zhao, Boya Wu, Shitao Xiao, Xi Yang, Yongping Xiong, Bo Zhang, Tiejun Huang, and Zheng Liu.
\newblock Mlvu: A comprehensive benchmark for multi-task long video understanding.
\newblock \emph{arXiv preprint arXiv:2406.04264}, 2024.

\end{thebibliography}
}

\end{document}